\documentclass{jdmdh}
\usepackage[utf8]{inputenc}
\usepackage{array}
\usepackage{pgfplots}
\usepackage{float}
\usepackage{tabularx}
\usepackage{titlesec}

\pgfplotsset{compat=1.18}

\titlespacing*{\section}
{0pt}{2.5ex plus 1ex minus .2ex}{1.3ex plus .2ex}
\titlespacing*{\subsection}
{0pt}{1.0ex plus 1ex minus .2ex}{1.3ex plus .2ex}
\titlespacing*{\subsubsection}
{0pt}{1.0ex plus 1ex minus .2ex}{1.3ex plus .2ex}

\newcolumntype{C}{>{\arraybackslash}X} 

\title{You Actually Look Twice At it (YALTAi): using an object detection approach instead of region segmentation within the Kraken engine}
\author[1]{Thibault Clérice}
\affil[1]{Centre Jean-Mabillon, PSL Research University, France} 
\affil[2]{Hisoma (UMR 5189), Université Lyon 3, Université de Lyon (UDL), France} 

\corrauthor{Thibault Clérice}{thibault.clerice@chartes.psl.eu}

\begin{document}

\maketitle

\abstract{Layout Analysis (the identification of zones and their classification) and line segmentation are the first steps in Optical Character Recognition and similar tasks. The ability of identifying the main body of text from marginal text or running titles makes the difference between extracting the full text of a digitized book and noisy outputs. We show that most segmenters focus on pixel classification and that polygonization of this output has not been used as a target for the latest competitions on historical documents (ICDAR 2017 and onwards), despite being the focus in the early 2010s. We suggest that transitioning the task from pixel classification-based polygonization to object detection using isothetic rectangles might improve results in terms of speed and accuracy. We compare the output of Kraken and YOLOv5 in terms of segmentation and show that the latter severely outperforms the first on small datasets (1110 samples and below). We release two datasets for training and evaluation on historical documents as well as a new package, YALTAi, which injects YOLOv5 in the segmentation pipeline of Kraken 4.1.}


\section{Introduction}

 
In recent years, automatic text extraction has become an important activity in digital philology and, in general, in corpus creation for historical documents. A few different tools share the software market: Transkribus and eScriptorium (with its underlying and autonomous Kraken engine) are probably the two best known platforms. As model recognition improves over the years, the question of quasi-automatic corpus extraction from digitized manuscripts and early printed material has become more of a question of appropriate tools rather than HTR performances for specific fonts or languages and periods. On the side of Kraken, which has the advantage of being fully open-source, unlike Transkribus, the layout segmentation \citep{kiessling2019badam} has suffered from poor performances user-perception, specifically with small dataset (around a thousand samples for train/dev/test) or very small dataset (less than 300 for train and dev). 

This results perception has slowed the improvement of automatic information extraction, where the main body of text could be drawn from a printed book or a manuscript by ignoring marginal texts such as footnotes or running title. If the segmenter is not able to recognize columns (or separate them) in medieval manuscripts or does not distinguish main bodies of text from foot notes, the co-occurrences of the extracted text are fooled by a wrong interpretation of the layout. Extracting main body of works could result in creating data minable corpora of unprecedented sizes. We know of three research projects whose goals are hindered by the inability of Kraken to be trained with small datasets:

\begin{itemize}
    \item Pierre-Carl Langlais, Jean-Baptiste Camps, Nicolas Baumard and Olivier Morin's work on the evolution of French literary fictions from 1050 to the 1920s;
    \item Ariane Pinche's research project on medieval vernacular \textit{Légendiers}, compilations of Saint's Lives in Old French, with manuscripts split in multiple columns;
    \item Simon Gabay and Béatrice Joyeux-Prunelle's work on manuscripts sales and art exhibition catalogues where the identification of each entry in the dataset leads to a lower amount of data to parse through natural language processing for denoising.
\end{itemize}

Layout Analysis has been one of the most frequent tasks in competition in conferences such as \textit{International Conference on Document Analysis and Recognition} (ICDAR) and the International Workshop on Historical Document Imaging and Processing (HIP). However, historical document layout analysis competition saw the light of the day in 2011, as a joint venture from ICDAR2011 and HIP2011. The 2011 event limited itself to print documents and focused on both the region segmentation as well as the region classification operations. In the context of ICDAR2011's Historical Document Layout Analysis competition (HDLA2011) and in general, layout segmentation is understood as the operation of recognizing blocks of content from the background in a digitized document. Region classification, on the other hand, is understood as the qualification of the blocks found during layout segmentation into various different categories. During ICDAR2011, documents for the competition spanned from the 17\textsuperscript{th} to the early 20\textsuperscript{th} century and were segmented using polygons, which is the norm for this competition.

In 2011, methods for segmentation would mostly use a binarization step followed by some form of separator or content detection. 
ICDAR2017 had three relevant competitions:
\begin{itemize}
    \item the \textit{competition on Page Object Detection} (POD2017), whose dataset is composed of scientific papers in which the objective is to detect figures, tables, equations, etc.;
    \item the \textit{competition on Historical Book Analysis} (HBA2017), which has been renewed in 2019 (HBA2019);
    \item the \textit{competition on Layout Analysis for Challenging Medieval Manuscripts} (HisDoc-Layout-Comp).
\end{itemize}

While POD2017 sets the objective as a bounding box detection task, both HBA and HisDoc-Layout-Comp treats the task as a pixel labelling one, a pixel can be assigned to multiple classes. Since 2017, deep learning models have emerged in the competition, specifically in HBA and HisDoc-Layout-Comp where CNN models shined. 

However, the shift from polygon detection at ICDAR2011 to pixel-based categorization at ICDAR 2019 creates a bias for the downstream task such as main body extraction: pixel categorization ignores the necessary serialization and clustering of said pixels into various zones that can then be filled with lines and content. This approach is specifically damageful in tools such as Kraken (v4.1.2) as it led to the creation of ``hacks'', which zones being qualified with \textit{ColumnEven} and \textit{ColumnOdd} systems in order to ensure the differentiation of columns in the clustering step\footnote{While undocumented, this was the approach taken by many projects including LECTAUREP under the advices of the SCRIPTA team who is working on Kraken and eScriptorium.}. Note that even stronger models such as ARU-NET \citep{gruning2018arunet} only computes pixel classification and does not polygonize the output in its official implementation\footnote{\url{https://github.com/TobiasGruening/ARU-Net}}.

With the specific purpose of training models on small datasets and document text extraction, we propose to approach the document segmentation as an object detection task, following the bounding box approach of ICDAR2017 instead of the polygon detection of ICDAR2011 or the pixel categorization of HBA or HisDoc-Layout-Comp. In this context, we propose two new datasets, a historical tabular document dataset (YALTAi-Tables \citep{clerice_thibault_2022_6827706}) and a manuscripts and early printed book dataset (YALTAi-MSS-EPB \citep{clerice2022yaltaimss}). We do not propose a specific model but instead reuse You Only Look Once (YOLOv5)'s tools and models. Results show up to 100 times better results on column detection for tabular documents as well as double the score of main body document detection. As YOLOv5 base version is limited to ``straight'' boxes, it limits the application to lightly or unskewed documents. Bounding box can also overlap easily and we lose the precision of polygons.

\begin{figure}[ht]
    \centering
    \includegraphics[width=\linewidth]{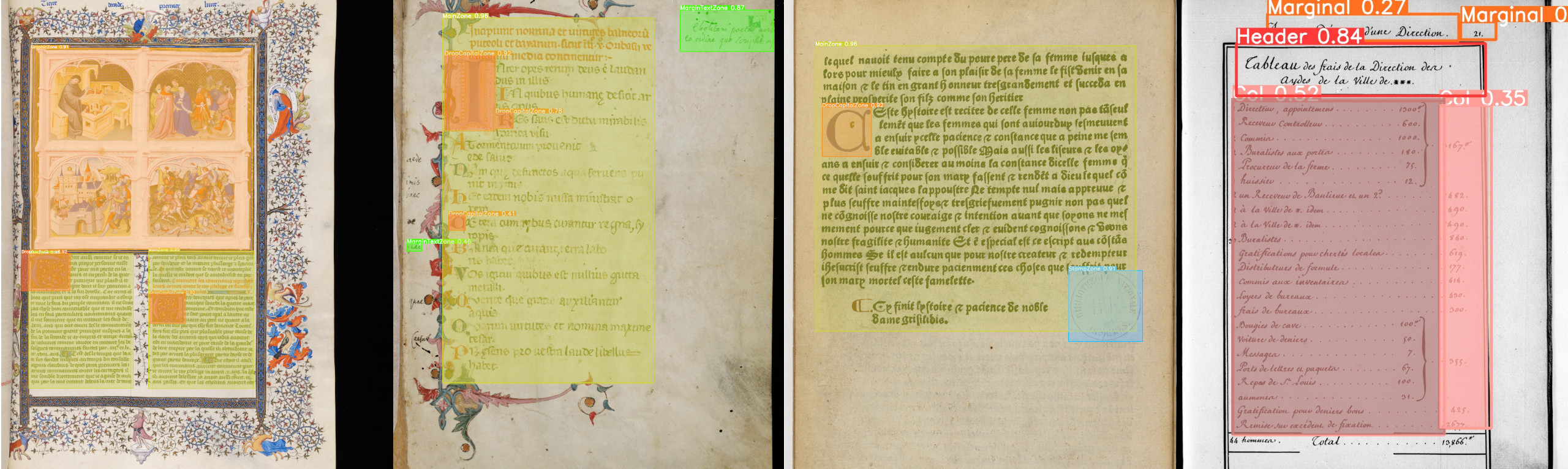}
    \caption{Prediction on the test set with YOLOv5x models for the Segmonto dataset (three pictures on the left) and the tabular dataset (last picture, columns are in alternating colours for readability). Illustrations are in orange (first picture top), drop capitals are in darker orange, marginal text in green, yellow is the main body of text.}
    \label{fig:4images}
\end{figure}

In summary, the contributions of this paper are

\begin{enumerate}
    \item a proposal for shifting from polygon and pixel labelling to bounding box detection for documents that permit it at the layout detection stage (including a shift regarding evaluation of said results from pixel classification to object detection);
    \item a new dataset for historical tabular documents, spanning from the 16\textsuperscript{th} to the early 20\textsuperscript{th} century;
    \item a new dataset for the segmentation of historical documents such as manuscripts and early printed books (from the 9\textsuperscript{th} century to the 16\textsuperscript{th}) following the Segmonto ontology for segmentation classification labels;
    \item models for the detection of content regions according to both the Segmonto guidelines and a tabular approach;
    \item a tool, YALTAi, which allows for  \begin{itemize}
        \item the conversion of ALTO to YOLOv5 formats and vice versa;
        \item accessing a command line interface similar to Kraken interface which injects the region detection of YOLOv5 before using the line serialization of Kraken\footnote{And thus, looking twice at the picture, once from YOLOv5, and a second time from Kraken line segmenter.}
    \end{itemize}
\end{enumerate}

\section{Background and Related Work}
\label{sec:related}

\subsection{Background}

On a printed or manuscript page, conventions lead the organization of content on a page, such that element of a document can easily be processed by a human reader: a letter is usually formed of an opener, a body with one or many paragraphs, and a closer; a printed book usually contains a main body of texts -- be it paragraphs or lines and poems -- and a running title, a page number and optionally some marginal information such as footnotes. Literary manuscripts follow such an organization, but are generally composed of two or more columns of main body of texts per page, each column is read from left to right so that when Column A is finished we start reading Column B. Such an organization is crucial for the understanding of the content, as it allows the identification of contiguous blocks of main text and blocks of additional information for the human reader.

Segmonto provides a welcome ontology and syntax for specifying the organization of pages from a layout perspective. \citet{gabay2021segmonto} define a series of top-level graphical structures that can be found in digitized document from medieval manuscripts up to contemporaneous documents: main bodies of text are qualified as \texttt{MainZone}, marginal text such as footnotes and scribbles as \texttt{MarginZone}, paintings, photos and figures in general are represented by \texttt{GraphicZone} and specific medieval and modern early printed book features such as illuminated capitals fall under the category of \texttt{DropCapital}. These categorizations can be refined at a second level with project-specific vocabulary which can contain categorization such as \texttt{MainZone:Entry} for a dictionary entry or \texttt{MarginText:Footnote}. Documents automatically or manually tagged using this ontology allows for text extraction and data mining at scale without worrying with the noise of running titles or marginal text: preliminary work by \citet{christensen2022gallic} shows this possibility.

In his article ``Where is digital philology going''\footnote{French: ``Où va la philologie numérique ?''.}, \citet{camps2018o} analysed that Handwritten Text Recognition (HTR) can now be placed upstream of philological study, with downstream research taking advantage of the mass of data and multi-scale analysis for different tasks, from quantitative palaeography to computational stylistics. This inclusion of HTR in the workflow of medievalist and philologists in general has profited from the development of easy-to-use, highly available and performant HTR and OCR user interface. Research infrastructure software such as Transkribus \citep{kahle2017transkribus} or eScriptorium have taken their place in the humanist's office.

\citet{kiessling2019escriptorium} present Kraken and eScriptorium. Unlike Transkribus, it is built around fully installable and open-source practices, allowing for a take-and-go approach and not making the user prisoner within an ecosystem. Kraken is the computational software for training, evaluation and inference of models for both segmentation and text recognition tasks. Before eScriptorium, it could be used only through its python API or most likely with its command-line interface (CLI). eScriptorium brought a user interface that allows for importing, segmenting, annotating and transcribing document, manually (no prediction provided by Kraken), semi-automatically (prediction provided and manually corrected) or automatically (prediction only).

\begin{figure}[h]
    \centering
    \includegraphics[width=\linewidth]{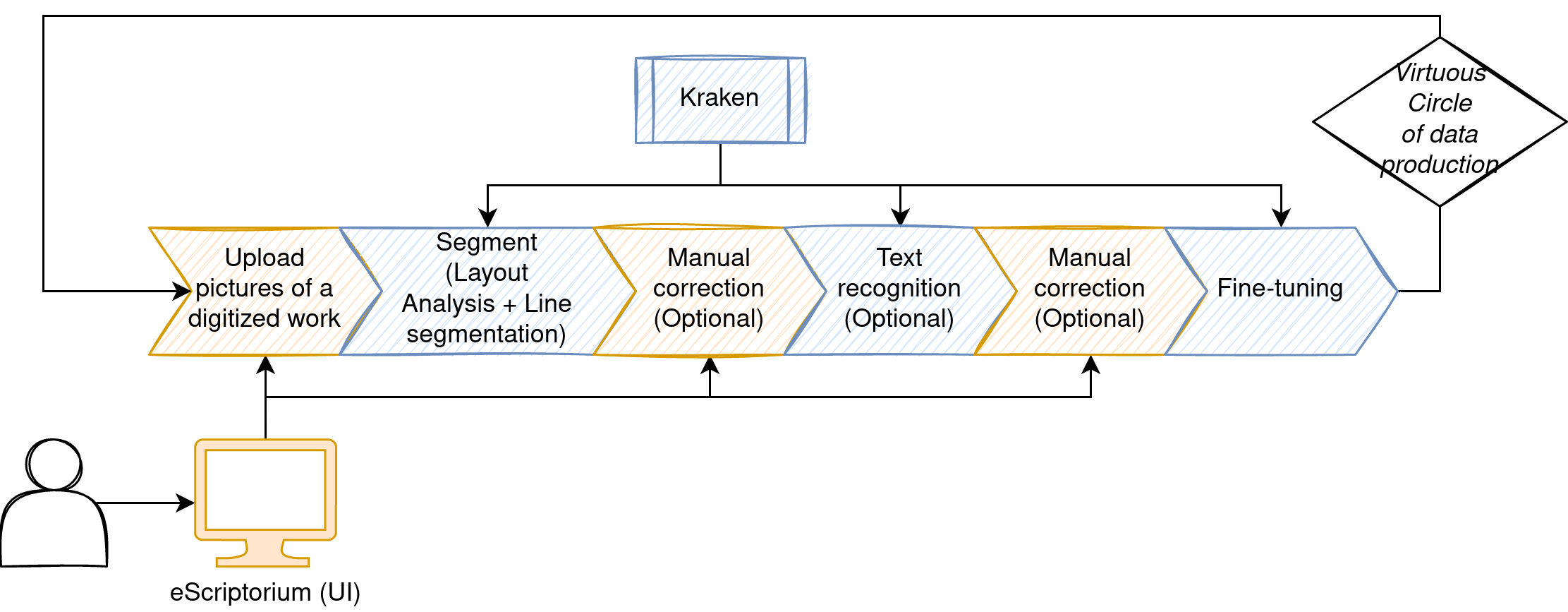}
    \caption{Typical Handwritten Text Recognition (HTR) workflow. A user uploads a set of pictures from a digitized book, segments the document both at the level of the layout and the lines, corrects the segmentation, provides a transcription or correct an automatic one and then fine-tune or creates models for other pages of the same document or documents of the same kind.}
    \label{fig:htr-workflow}
\end{figure}

\subsection{Related Works}

In HDLA2011, \citet{antonacopoulos2011historical} defines layout analysis as both the actions of identifying ``printed regions (Page segmentation) and labelling them according to the type of their content (Region Classification)''. Competitions of page segmentation have been run at ICDAR since 2001 and mostly on contemporary documents. 2011 was the first historical document oriented competition but restricted itself to printed material up to the 17\textsuperscript{th} century. In \citet{antonacopoulos2009icdar}'s review of ICDAR2009 Page Segmentation competition (PS2009), ground truth is defined by ``isothetic polygons (i.e. a polygon having only horizontal and vertical edges) [as] such a representation enables a very accurate and efficient geometric description, especially for complex-shaped regions''. 

In the late 2000s and early 2010s, competitions were dominated by difficulties such as binarisation, recognition of separators (whether background as white space or printed such as column borders), combined with different approaches to image analysis. As of ICDAR2017, Convolutional Neural Networks and various other deep learning methods~\citep{hashmi2021castabdetectors, pfitzmann2022doclaynet} appeared on this subject, in HBA and HisDoc-Comp-Layout or in unrelated settings. However, \citet{antonacopoulos2009icdar} showed that non-deep learning methods still gave good results on complex document layout with DSPH getting the highest score of the competition: \citet{lu2021probabilistic} proposes a probabilistic approach to document segmentation and beats the first deep-learning-based method of the competition by 15 points (95.1\% vs. 79.6\%).

In parallel with the development of layout analysis research, object detection research has made great strides from still image application to extending performance to real-time detection. Contrary to layout analysis, the object detection task aims at detecting so-called objects in pictures such as dogs, people or boats. It mainly operates with ``straight'' bounding boxes (with sides of the bounding box being only horizontal and vertical lines) but it also exists with oriented bounding boxes or masks. Unlike layout analysis, it deals with three-dimensional information and as such different scales and different layers of information, from the foreground to the background. One of the main datasets for object detection model evaluation is the Microsoft COCO dataset and its evolutions. Described in \citet{lin2014microsoft}, COCO contains 91 kinds of objects and 2.5 million instances across 328,000 images.

In 2015, \textit{You Only Look Once} (YOLO) was released with its version 1 (YOLOv1). Each YOLO model has had in its core goals to detect quickly and close to real-time instances of objects are at a high success rate. YOLOv1 used 24 convolution layers with two fully connected layers, splitting the picture into multiple cell and predicting for each cell what it could contain \citep{jiang2022a}. The use of YOLOv5 \citep{jocher2022ultralytics} or its predecessors in combination with OCR task is not novel~\citep{huang2019yolo}, specifically for text transcription in scenes (\textit{e.g., }panels or license plate \citep{raj2022license}) or highly formatted documents such as receipts \citep{lin2022automatic}. Similarly to our work, \citet{ning2021mt} uses YOLOv5 on modern and contemporary documents for table detection but limits its work to table instances with scores on the lower end of the spectrum of state-of-the-art (SOTA) methods but with significantly fewer parameters (240M parameters for the best score vs. 7M for the implemented solution).


\section{Method}

We propose to shift the layout analysis task from a pixel categorization or polygon segmentation task to an object detection task, where each region is an instance of a category of zone, mixing the segmentation and classification part of the layout analysis. For the purpose of this paper, we work only with isothetic rectangles which are produced by using the minimal and maximal $X$ and $Y$ of each polygon, which results in a more important area captured by each zone. 

\begin{figure}
    \centering
    \includegraphics[width=.5\linewidth]{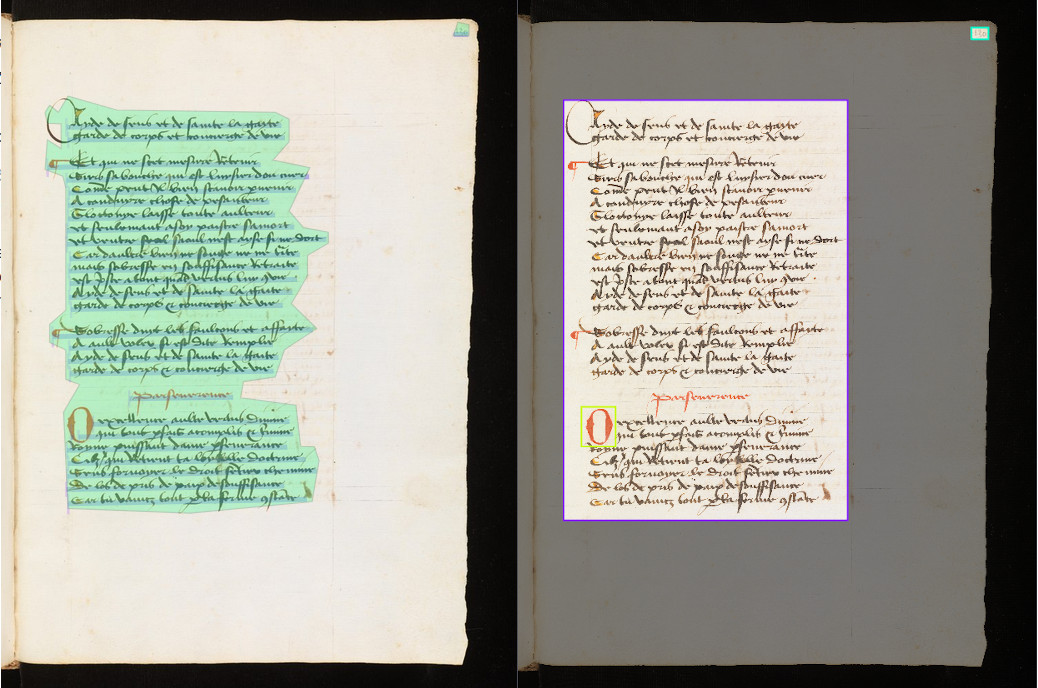}
    \caption{Example of polygon in the ground truth when the Kraken prediction is correct on the left. On the right, its simplification into an isothetic rectangle for object detection.}
    \label{fig:yaltai:rectangulisation}
\end{figure}

In terms of software and model training, each model is kept independent: YOLOv5 is trained within the YOLOv5 framework and can profit from its ecosystem while Kraken line segmentation is also trained separately. Both models are then used together at inference or evaluation time.

\begin{figure}
    \centering
    \includegraphics[width=.8\linewidth]{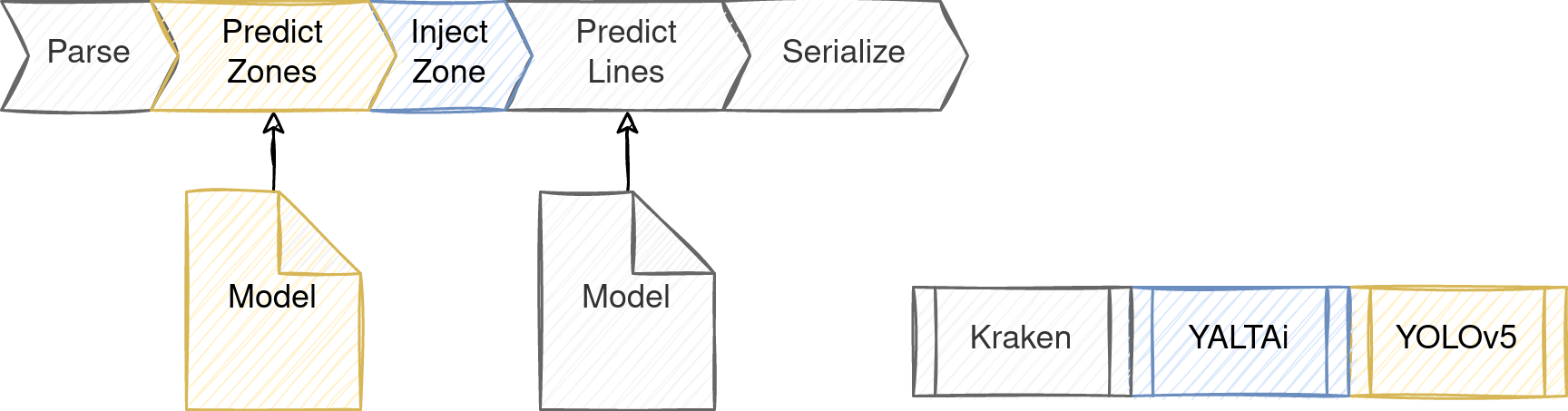}
    \caption{Workflow and responsibilities at inference time.}
    \label{fig:yaltai:injection}
\end{figure}

As the objective is to provide end users with a readily usable analysis of documents, the segmentation of zones should be fed in Kraken before the line detection is serialized, in order to dispatch lines into the various regions detected by YOLOv5(see Figure \ref{fig:yaltai:injection} for the details of the workflow at prediction time). In the YALTAi software library, YOLOv5's output is serialized in the same fashion as Kraken region's segmentation output. Using Kraken own region-line handler\footnote{Kraken uses the interpolation of the baseline to find the middle point and then uses the presence of the calculated point inside the region polygon to decide whether it belongs to the region.}, it then distributes lines segmented by Kraken into the YOLOv5 precomputed regions.

\section{Experimental Setup}

The experimental setup is focused on evaluating the gains in terms of accuracy of zone detection or segmentation in historical documents with ``complex'' layout. We compare YOLOv5 models to Kraken segmentation model using the same datasets (described below) with fix splits\footnote{Splits are available in the dataset published with the present article.} and using Mean Average Precision (mAP) as well as Average Precision (AP) for each class of categories that are relevant to the qualitative analysis of results.

\subsection{Datasets}

\subsubsection{Manuscripts and Early Printed Books}

We built two datasets for the experiment and both Kraken and YOLOv5 are trained on the same rectangle-converted dataset. The first dataset consists of manuscripts and early printed books (YALTAi-MSS-EPB) from the 9\textsuperscript{th} century up to the 17\textsuperscript{th} century. It is based on five datasets:

\begin{itemize}
    \item CREMMA Medieval \citep{pinche2022cremma} focused on manuscripts in Old French with various layout and the inclusion of microfilm digitizations (microfilm being black and white pictures).
    \item CREMMA Medieval LAT \citep{clerice2022cremma} contains manuscripts in Latin with different types of content, from poetry to medical content, which provides a diversity of layouts.
    \item Eutyches \citep{vlachou-efstathiou2022voss} focused on two manuscripts from the Early Middle Ages.
    \item Two datasets from the Gallicorpora project (\citep{pinche2022mss, pinche2022goth}) containing either incunabula, printed books with gothic scripts or manuscripts from the 15\textsuperscript{th} centuries, all written in Old or Middle French.
\end{itemize}

On top of these five datasets, we produced an addendum of 593 annotated images for this paper to reach the objective of 1,000 images. This additional dataset provides a higher number of samples (123 different documents) and spans from the Middle Ages to the early modern era (17\textsuperscript{th} century). These new annotated pages were produced with early versions of the model and corrected by hand on the Roboflow application, thus they are more specific to the way YOLOv5 detects images and less prone to over-estimated area of polygons based on rectangle simplification as they are manually tuned to the images. The whole dataset uses Zone types from Segmonto, thus avoiding the project-specific and vocabulary free levels of subtypes and numbers that come with Segmonto deeper levels. 

\begin{table}[h]
    \centering
\resizebox{\linewidth}{!}{
    \begin{tabular}{l|rrrrrr}
         \hline
         Dataset & ``Books'' & Pages & Starting century & Ending century & Type & With B\&W \\ \hline
         \citep{pinche2022cremma} & 13 & 263 & 12 & 15 & Manuscripts & Yes \\
         \citep{clerice2022cremma} & 7 & 30 & 13 & 15 & Manuscripts & Yes \\ 
         \citep{vlachou-efstathiou2022voss} & 2 & 129 & 9 & 9 & Manuscripts & No \\
         \citep{pinche2022mss} & 1 & 20 & 15 & 15 & Manuscripts & No \\
         \citep{pinche2022goth} & 4 & 80 & 16 & 16 & Printed & No \\
         \textit{Original Data} & 123 & 593 & 9 & 17 & Mixed & No \\ \hline
         \textbf{Total} & 150 & 1110 & 9 & 17 & Mixed & Yes \\ \hline
    \end{tabular}%
    }
    \caption{Quantitative and qualitative description of YALTAi MSS EPB. ``Books'' can be full manuscripts or printed books. \textit{Original data} are previously unpublished data produced directly using bounding boxes.}
    \label{tab:dataset:mssepb}
\end{table}

Data were mixed semirandomly into 3 split for training (854 images) and development (154), with works being able to be in two or more splits. Test (139) is only composed of different sources. The MainZone bounding box are the largest areas of all zones and just behind the DropCapitalZone in terms of quantity. On the other hand, DigitizationArtefactZone and DamageZone are overly represented in the test set (see Table \ref{tab:comp:segmonto} for the statistical details).

\begin{table}[ht]
    \centering
    \begin{tabular}{l|rrr|r|rr}
    \hline
                              &   Train &   Dev &   Test &   Total &   Average area &   Median area \\
    \hline
     DropCapitalZone          &    1537 &   180 &    222 &    1939 &           0.45 &          0.26 \\
     MainZone                 &    1408 &   253 &    258 &    1919 &          28.86 &         26.43 \\
     NumberingZone            &     421 &    57 &     76 &     554 &           0.18 &          0.14 \\
     MarginTextZone           &     396 &    59 &     49 &     504 &           1.19 &          0.52 \\
     GraphicZone              &     289 &    54 &     50 &     393 &           8.56 &          4.31 \\
     MusicZone                &     237 &    71 &      0 &     308 &           1.22 &          1.09 \\
     RunningTitleZone         &     137 &    25 &     18 &     180 &           0.95 &          0.84 \\
     QuireMarksZone           &      65 &    18 &      9 &      92 &           0.25 &          0.21 \\
     StampZone                &      85 &     5 &      1 &      91 &           1.69 &          1.14 \\
     DigitizationArtefactZone &       1 &     0 &     32 &      33 &           2.89 &          2.79 \\
     DamageZone               &       6 &     1 &     14 &      21 &           1.50 &          0.02 \\
     TitlePageZone            &       4 &     0 &      1 &       5 &          48.27 &         63.39 \\
    \hline
    \end{tabular}
    \caption{Distribution of objects instances across datasets splits after random shuffling of pages for the YALTAi-MSS-EPB dataset.}
    \label{tab:comp:segmonto}
\end{table}

\subsubsection{Tabular dataset}
\label{subsubsection:tabulardataset}

\begin{figure}
    \centering
    \includegraphics[width=\linewidth]{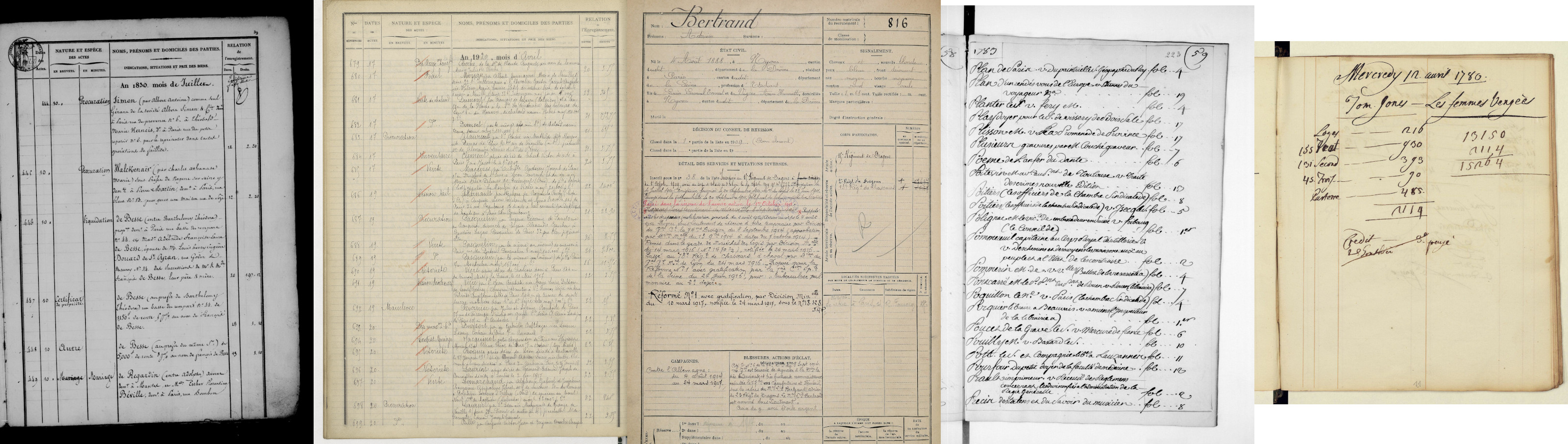}
    \caption{Tabular dataset excerpts. Two images on the left are from the Lectaurep dataset, three images on the right are original data for the test set.}
    \label{fig:dataset:table}
\end{figure}

The tabular dataset was produced using a single source, the \textit{Lectaurep Repertoires} dataset \citep{p2021notaires}, which served as a basis for only the training and development split. The test set is composed of original data, from various documents, from the 17\textsuperscript{th} century up to the early 20\textsuperscript{th} with a single soldier war report. In contrast to the MSS-EPB test set, the test set for tabular datasets falls entirely out of the domain. It presents situations where column borders are neither drawn nor printed, adopting a masonry layout—particularly evident in the soldier war report (refer to Figure \ref{fig:dataset:table})..

The tabular dataset contains four kinds of zones: Col (Columns), Header, Marginal, Text. The test set was only annotated with Columns and Header, as marginal and text were not clearly defined. The tabular dataset was simplified from the original dataset from \citet{p2021notaires}, as it used a known hack to work ``as intended'' with Kraken. We knowingly put some stress on Kraken on this test. As Kraken relies on pixel classification and less on polygon or bounding box detection, it usually merges together regions of the same kind that are too close to each other. To reduce the effect of these merging errors, the trick learned by the community is to split zones into alternating names, such as ColOdd and ColEven. In the Lectaurep dataset, we found around 16 different ways to describe columns, from Col1 to Col7, the case-different col1-col7 and finally ColPair and ColOdd, which we all reduced to Col (see Table \ref{tab:comp:tabular} for the statistical details).

\begin{table}[ht]
    \centering
    \begin{tabular}{l|rrr|r|rr}
    \hline
              &   Train &   Dev &   Test &   Total &   Average area &   Median area \\
    \hline
     Col      &     724 &   105 &    829 &    1658 &           9.32 &          6.33 \\
     Header   &     103 &    15 &     42 &     160 &           6.78 &          7.10 \\
     Marginal &      60 &     8 &      0 &      68 &           0.70 &          0.71 \\
     Text     &      13 &     5 &      0 &      18 &           0.01 &          0.00 \\
    \hline
    \end{tabular}
    \caption{Annotation and details about the annotations across splits of the tabular  dataset.}
    \label{tab:comp:tabular}
\end{table}

\subsection{Model Training}

Kraken models were trained according to two different configurations, one with an input resize of 1200 and a second with an input resize of 640, like the recommended YOLOv5 input size\footnote{However, by its very nature, Kraken's algorithm is expected to suffer from lower resolution: comparison is here to provide an insight regarding image processing time and compute footprint in general rather than scores.}. Although Kraken recommends since recently a 1800 resized input, the training, with shuffled input, crashed many times because of the GPU RAM limits (roughly 11 GB). The model was trained for 50 epochs and the best model was selected according to mAP scoring on the development set. Scoring with mAP was produced outside of Kraken as the latter does not provide a best model selection. 

YOLOv5 models were trained with the latest commit (29d79a6) from the 2\textsuperscript{nd} of July. For both datasets, a model was trained using a YOLOv5x weights for initialization. A second YOLOv5n model was trained for the YALTAi-MSS-EPB dataset to evaluate the impact of the reduction of the size of the models on results: YOLOv5x model is advertised as a 166 MiB model while YOLOv5n weights only 4 MiB. Both models were trained during 50 epochs. The best model for evaluation was chosen according to YOLOv5 appointed best model.

Models were trained on a Nvidia RTX2080TI with 11 GB of RAM and a TDP of 250 W, on Ubuntu with CUDA 10.2 on drivers 440.118.02. Training for both tools used the same dataset with boundinges box coordinates rather than complex polygon.

\subsection{Evaluation}

The evaluation is done based on mean average precision (mAP) and average precision (AP). For fairness of the evaluation and to avoid any discrepancies in calculation framework, we used \texttt{mean-average-precision==2021.4.26.0} on the converted ALTO or the YOLOv5 text format. mAP is used with an intersection over union threshold of 0.5 (usually noted mAP@.5). Evaluation for Kraken used bounding box converted from the complex polygons it predicted.

\section{Experiments}

\subsection{Results for YALTAi MSS-EPB}

On the YALTAi MSS-EPB, the mAP of Kraken (6.95\%) results is largely beaten by the YOLOv5x (47.75\%) and the YOLOv5n model (34.63\%, see Table \ref{tab:scores:segmonto}). This lower results is due in part to lower results for the MainZone and DropCapital recognition, which are the most common objects in all splits (YOLOv5n, the worst-performing YOLOv5 model, doubles the performance of Kraken on MainZone and nearly triples them on DropCapital). On the other hand, Kraken is also unable to beat a 0\% AP on any of the less frequent objects such as MarginText, Numbering, QuireMarks or RunningTitle, lowering its capacity to reach a higher mAP. mAP being a macro-average metric, it favours models performing on all classes rather than a few classes. However, the score discrepancy on the MainZone is important enough that it wagers using YOLOv5 instead of Kraken segmenter for extracting main body of texts.

As for the difference between the nano (YOLOv5n) and the extra-large (YOLOv5x) models, the difference is small enough on the most frequent zone to accept a performance drop for a faster computation time. However, the YOLOv5n is largely beaten on less frequent classes such as MarginTextZone or NumberingZone.

\subsection{Results for YALTAi-Table}

\begin{figure}[h]
    \begin{minipage}{.5\linewidth}
        \centering
        \resizebox{.75\linewidth}{!}{%
            \begin{tabular}{l|r|rrrr}
            \hline
                    & mAP           & Col           & Header       \\ \hline
            Kraken  & 0.09          & 0.1           & 0.1          \\
            YOLOv5x & \textbf{4.77} & \textbf{12.9} & \textbf{1.4} \\  \hline       
            \end{tabular}%
        }
    \end{minipage}%
    \begin{minipage}{.5\linewidth}
        \includegraphics[width=\linewidth]{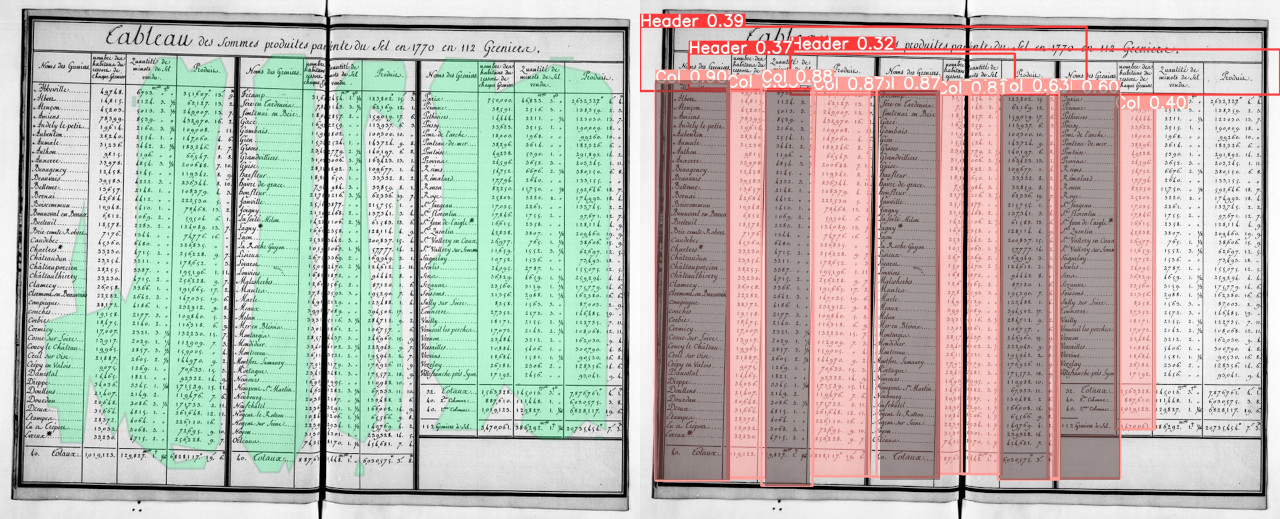}
        \corrauthor{}{}
    \end{minipage}%
\caption{On the left, scores of Kraken and YOLOv5 best models for the Table Dataset. \textit{Col} and \textit{Header} are the only annotated zones in the test set. Scores for zones are the Average Precision. Both Mean Average Precision and Average Precision are given in percents. On the right, two examples of predictions (Kraken on the left, YOLOv5 on the right): we see that Kraken does not separate columns and merges up to four columns while YOLOv5x fails to see only two columns but splits correctly the zones. See \ref{subsubsection:tabulardataset} for more explanation on the difficulties of the Kraken segmenter.}
\label{tab:scores:table}
\end{figure}

The results on YALTAi table are somewhat unfair to the Kraken segmenter but shows the impact of the object detection approach on a segmentation task. As seen in the example output at Figure \ref{tab:scores:table}, Kraken is unable to distinguish multiple columns and often merges the various detected zones. 

On the other hand, YOLOv5x shows a very high capacity to extend to different kinds of inputs as the input tabular document is not from the LECTAUREP dataset nor is the same in terms of content. At best, the document is also a printed document with various columns filled with manuscript content. In terms of genericity and ability to learn, the test results show that YOLO can learn from very few documents compared to Kraken, as it misses only a few columns.

\subsection{Efficiency}

We monitored the GPU usage in terms of memory, time and peak power consumption during the training of all models (see Table \ref{tab:scores:configs}). On the same system, YOLOv5 has shown a better efficiency than Kraken for each model and metric except for Kraken on the tabular dataset and power consumption. The training time seems to scale with the batch size but we are unable to confirm it, as Kraken does not allow to use batches for training. The inference time was only reported for YOLOv5 models to show the overhead that it would cost per image prediction: YOLOv5 has been designed for quasi-real time prediction on video streams, which explains the very low inference time.

\begin{table}[ht]
\resizebox{\linewidth}{!}{
\begin{tabular}{l|r|rrrrrrrrr}
\hline
        & mAP            & Main          & Graphic        & DropCapital     & MarginText     & Numbering      & QuireMarks     & RunningTitle   & Stamp        \\ \hline
Kraken  & 6.98           & 43.5          & 16.1           & 23.3            & 0.0            & 0.0            & 0              & 0              & 0            \\
YOLOv5x & \textbf{47.75} & \textbf{91.7} & \textbf{48.4}  & \textbf{69.2}   &  \textbf{48.3} &  \textbf{75.8} & \textbf{46.3}  &          45.6  & \textbf{100} \\ 
YOLOv5n &         34.63  &        87.0   &         44.2   &         54.6    &         13.9   &         43.0   &         25.0   &  \textbf{48.1} & \textbf{100} \\ \hline
\end{tabular}%
}
\caption{Scores of Kraken and YOLOv5 best models for the Segmonto MSS and Early Printed Books Dataset. Zones (Main, Graphic, Drop Capital, Margin Text, Numbering, Quire Marks, Running Title and Stamp) are the most important zones from the test dataset. Scores for zones are the Average Precision. Both Mean Average Precision and Average Precision are given in percents. Surprisingly, YOLOv5n beats YOLOv5x on RunningTitle but is most often largely outperformed on other zones. Kraken is completely outperformed with YOLOv5x more than doubling its scores on the main body zones, and roughly multiplying it by 7 times for the mean average precision.}
\label{tab:scores:segmonto}
\end{table}

\begin{table}[H]
\resizebox{\linewidth}{!}{
\begin{tabular}{l|rrrrrrrr}
\hline
       & Dataset  & Batch Size & Architecture & Size (MiB)      & GPU Power (Max W) & GPU RAM \% (Max)   & Training Time   & Prediction Time \\ \hline
Kraken & Table    & 1          & 640          & \textbf{5}      & 202.1             & \textbf{25.44 \%}  &  00:50          & -               \\
YOLOv5 & Table    & 2          & 5x           & 175             & \textbf{157.2  }  & 67.85              &  \textbf{00:25} & -               \\\hline
Kraken & Segmonto & 1          & 1200         &         5       & 236.1             & 66.02 \%           &  06:27          & -               \\
YOLOv5 & Segmonto & 2          & 5x           & 175             & 160.43            & 39.31 \%           &  \textbf{03:04} & 0.025s          \\
YOLOv5 & Segmonto & 2          & 5n           & \textbf{3.9}    & \textbf{54.05  }  & \textbf{11.5 \%}   &  03:11          & \textbf{0.004s} \\ \hline
\end{tabular}
}
\caption{Other metrics related to Kraken and YOLOv5. YOLOv5x produces the heaviest model (175mb, 35 times the weight of the Kraken model) but uses at most 77\% of the maximum power draw of Kraken. Training time is generally doubled under Kraken but corresponds roughly to the batch size division. Prediction time is per image. For the Segmonto model, the maximum GPU RAM usage is halved for YOLOv5n, while using a batch of 2, but is nearly three times the footprint of Kraken for the Table dataset. Training and prediction occured on a NVIDIA RTX2080TI with a rated 250W of powers and 11,018MiB of RAM on drivers 440.118.02 and CUDA 10.2.}
\label{tab:scores:configs}
\end{table}

\section{Conclusions}

Handwritten Text Recognition or Optical Character Recognition are the second step in text acquisition, right after layout analysis. However, in the main open source system used in the humanities, Kraken, the layout analysis under-performs severely on small datasets, specifically for the same-class regions that are close neighbours in the image, as these are often merged. For text extraction, identifying regions is of primordial importance as they can help filter out noise such as running titles or marginal addendum, for both manuscripts and printed books.

In this paper, we proposed to shift the layout analysis task to an object detection task, shifting in terms of predicted output from polygons to isothetic bounding box in the context of YOLOv5. We show that YOLOv5 severely outperforms Kraken on the same dataset for region segmentation, both in terms of evaluation metrics and GPU efficiency (Peak power usage and RAM).

The drawback of this approach lies in the bounding box limitations: bounding box limits us to regions that are rectangular and cannot cover complex polygons such as P- or C-shaped ones. We used YOLOv5 which limits us to isothetic bounding boxes but note that a YOLOv5 Oriented Bounding Box (YOLOv5-OBB) adaptation exists and could help for documents such as tables for a ``perfect'' coverage of each column without pre-processing and deskewing.

Taking advantage of the easy to use, open-sourced and openly licensed python API of Kraken, we are able to often a new package, \texttt{YALTAi}, which simply plugs in the YOLOv5 model before line dispatching across regions. It possess the same segmentation CLI interface as Kraken but adds a simple \texttt{--yolo} parameter that allows for selecting a YOLOv5 model for region segmentation.

\section*{Software}

Source code is available at \url{https://github.com/ponteineptique/YALTAi} and software can be installed using \texttt{pip install yaltai}.

\section*{Acknowledgements}

This work was made possible through the funding of the DIM MAP, now called DRIM PAMIR.

\bibliographystyle{plainnat}
\bibliography{article}

\end{document}